\documentclass[a4,12pt,english]{article}
\usepackage {algpseudocode}
\usepackage {algorithm}
\usepackage{longtable}
\usepackage{graphicx}
\usepackage{amssymb}
\begin{document}
\title{Superclustering by finding statistically significant separable groups of optimal gaussian clusters}
\author{Oleg I.Berngardt}

\maketitle

%% Text of abstract
\begin{abstract}
The paper presents the algorithm for clustering a data set by grouping
the optimal, from the point of view of the Bayesian information criterion, number of
Gaussian clusters into the optimal, from the point of view of their
statistical separability, superclusters.

The algorithm consists of three stages: representation of the data set
as a mixture of Gaussian distributions - clusters, which number is
determined based on the minimum of the Bayesian information criterion; using the Mahalanobis
distance to estimate the distances between the clusters and cluster
sizes; combining the resulting clusters into superclusters using the
density-based spatial clustering of applications with noise method 
by finding its maximum distance hyperparameter providing
maximum value of introduced matrix quality criterion at maximum number
of superclusters. The matrix quality criterion corresponds to the
proportion of statistically significant separated superclusters among
all found superclusters.

The algorithm has one hyperparameter - statistical significance
level, and automatically detects optimal number and shape of superclusters
based of statistical hypothesis testing approach. The algorithm demonstrates
a good results on test data sets in noise and noiseless situations.
An essential advantage of the algorithm is its ability to predict
correct supercluster for new data based on already trained clusterer
and perform soft clustering. The disadvantages of the algorithm
are: its low speed and stochastic nature of the final clustering.
It requires a sufficiently large data set for clustering, which is
typical for many statistical methods.

\end{abstract}

\section{Introduction}

Data clustering algorithms \cite{REVIEW_2017} are usually divided
into two big categories: hierarchical and partitional. Hierarchical
methods are based either on joining close clusters (agglomerative
methods) or separating big clusters (divisive methods). Partitinal
methods can be divided on distance based, model based and density
based. Model-based ones describe the data by a combination of statistical
distribution functions, for example EM algorithm for the mixture of
normal distributions \cite{Reynolds_2015}. The basis
of distance methods is the pairwise distance between points, for example
spectral clustering \cite{SPECTRAL}. The popular density method
is Density-based spatial clustering of applications with noise (DBSCAN),
suggested in \cite{DBSCAN_1996}, and discussed in \cite{DBSCAN2_2017}.
Comparison of different methods can be found in \cite{REVIEW_2018,REVIEW_2019}.

Most of the methods have one or more hyperparameters \cite{REVIEW_2019}
- for DBSCAN this is the maximum distance between a pair of objects,
for K-means\cite{KMeans}, for Spectral\cite{SPECTRAL} and Gaussian
Mixture, this is the number of classes. One of the main tasks for
a researcher is to choose a clustering method and find the hyperparameters
values that provide the best solution of his clustering problem.

The choice of a clusterer is related to the data model - the clusterer
that is optimal for the researcher should correspond to the data.
On the one hand, it should have the necessary freedom to correctly
describe the data set for any expert separation of the data set into
a clusters - in terms of the Kleinberg theorem \cite{Kleinberg_2002}
- to provide richness. On the other hand it should correctly label
the original unlabeled data set so that the number and shape of clusters
coincide with what an expert expect from the data set.

Let us explain the main idea of the algorithm, suggested in the paper.
A data set consisting of two expert-labeled classes can be separated
into an arbitrary number of clusters, and by many ways.

Let us consider the Rand index \cite{RAND} at data set $S$ in classification
(not clusterization) problem:

\begin{equation}
RI=\frac{TP+TN}{C_{2}^{N}}
\label{eq:RI}
\end{equation}

where TP (True Positive) is the number of pairs of elements that are
in the same clusters in set $S_{l}$ (data set $S$, labeled by an
expert) and in the same cluster in set $S_{c}$ (data set $S$, clustered
by an algorithm); TN (True Negative) is the number of pairs of elements
that are in different clusters in set $S_{l}$ and in different clusters
in set $S_{c}$; $C_{2}^{N}$ - total number of pairs in data set $S$.

In this case, pairwise true positive (PWTP) metric can be defined
as:

\begin{equation}
PWTP=\frac{TP}{C_{2}^{N}}
\label{eq:PWTP}
\end{equation}

and pairwise true negative (PWTN) metric can be defined as:

\begin{equation}
PWTN=\frac{TN}{C_{2}^{N}}
\label{eq:PWTN}
\end{equation}

Rand Index is sum of PWTP and PWTN. Obviously, the easiest way to
achieve the maximum of PWTN metric, is to put each point into its own separate
cluster, but PWTP metric in this variant will become zero. The easiest
way to achieve the maximum of PWTP metric, is to put all the points
into the same single cluster, but PWTN metric in this variant will
become zero.

Let us refer the optimal PWTN algorithm as the algorithm providing
maximum PWTN, and refer the optimal PWTP algorithm as the algorithm
providing maximum PWTP. We suggest that by corresponding grouping
clusters produced by optimal PWTN algorithm we can obtain optimal
PWTP algorithm without loosing its PWTN optimality.
Adequate classification can be defined as a classification
that is both optimally PWTP and optimally PWTN with respect to the
teacher's labeling. It is obvious that this corresponds to the absolute
maximum of RI because in this case the errors of the first and second
kind (FN, FP) are equal to zero. The absolute maximum of both PWTP
and PWTN metrics is reached simultaneously when the number of classes
in the labeled data set $S_{l}$ and classified data set $S_{c}$ are
the same, and their shapes are also the same. Therefore, the adequate
classification problem in the presence of a labeled data set $S_{l}$
can be considered as the problem of achieving RI its absolute maximum
value of 1. A good approximation to adequate classification can be
found by supervised learning, when $S_l$ is given.

However, in most cases, it seems to be impossible to achieve adequate
clustering by unsupervised learning. We do not know the set $S_{l}$,
and therefore RI will be most likely less than 1 due to the imperfection
of our knowledge about $S_{l}$. Therefore, each clustering model
has its own limitations - our initial assumptions about the unknown
set $S_{l}$: K-Means searches for centroids, DBSCAN for distance
separated points, Gaussian Mixture for points which distributions are
close to normal ones.

Approximation by Gaussian distributions is widely used in clustering
because potentially it can provide a high PWTN metric - an infinite
sum of Gaussian distributions can fit (and separate) any arbitrary
discrete set of points.

This paper demonstrates an algorithm that uses grouping of gaussian
clusters into superclusters to create a clustering close to adequate
one, based on our own assumptions about the set $S_{l}$, that will
be explained later.

\section{Algorithm}

\subsection{The idea}

We will solve the problem sequentially: at the first stage, we will
approximate our data most accurately by the optimal number of clusters
from the point of view of likelihood (we will construct a clustering
with a potentially high PWTN metric). At the second stage we will
group these clusters into the optimal number of superclusters (we
will increase PWTP metric) from the point of view of the distance
between these superclusters. Optimality in terms of likelihood at
the first stage should allow us to best separate the data clusters
from a statistical point of view, and at the second stage we group
too close clusters. The problem conceptually is close
to the well-known hierarchical method - agglomerative clustering\cite{Augmentation},
but uses only two stages instead of using their sequence.

Important requirements for our algorithm are: robustness to a noise,
interpretability of each of its stages in terms of a statistical approach,
and minimizing the number of hyperparameters.

The first stage can be simply solved when we are not limited by the
maximum number of clusters: we can choose the number of clusters equal
to the number of objects and place each object in its own cluster.
If we want to limit the number of clusters we need to use some kind
of clustering as an initial stage. We choose an approximation by mixture
of normal distributions - Gaussian Mixture (GM). The advantage of
this method is that it is statistical one. Another advantage - it
has a widely used criterion for finding the optimal number of objects
- the Bayesian Information Criterion - BIC \cite{BIC}. Therefore,
we expect that most likely the optimal number of initial clusters
will be less than the number of points (objects) in the data set. The
disadvantage of this method is its stochastic nature associated with
the iterative Expectation-Maximization (EM) algorithm used for search
of GM parameters, and dependence of EM results on its stochastic initial
state \cite{Reynolds_2015}.

The second stage is close to agglomerative clustering. The main principle
of agglomerative clustering is sequential grouping of the closest
clusters. In agglomerative approach, one should group them until all
the clusters are grouped into a single supercluster, or until necessary
number of superclusters is reached, or until optimum of a certain criterion
is reached.

Therefore two main problems arise: by what principle to combine clusters,
and by what criteria to detect if the clustering becomes the optimal
one. There are many methods for solving the first problem. The simplest
method is sequential one - greedy joining. In this case, we find a
pair of closest clusters, and group them into one \cite{Augmentation}.
Another common method is density based distribution (DBSCAN). However,
DBSCAN has problems in choosing value of a hyperparameter - the maximum
distance between points in a cluster ($\varepsilon$). So one need
to vary (or to use some search method) to find optimal $\varepsilon$,
which slows down the algorithm. We choose DBSCAN method due to it
is easy to implement.

The second problem is the choice of metric - the distance between
clusters. Since we use GM, one of the native metrics is the Mahalanobis
distance \cite{Mahalanobis}, which scales the space depending on
the covariance matrix of the found Gaussian distribution. The distance
between two clusters can also be calculated using in terms of this metric. Our
algorithm stops when the distance between all found superclusters
exceeds a given threshold level.
Mahalanobis distance is widely used in solving classification problems, 
for example in Quadratic Discriminant Anaysis \cite{McLachlan_1992}.

The name of our algorithm (GMSDB) is formed from the three elements
it based on: Gaussian Mixture (GM), statistical approach (S) and DBSCAN
(DB). Let us describe the GMSDB algorithm in details.

\subsection{Stage 1: Data approximation by the optimal GM}

At this stage, we iterate over the number of clusters $N$ until the
minimum of the BIC criterion\cite{BIC} is reached, or until the
number of clusters reaches a high enough value that we have specified.
In this stage, we assume that the set of points $\overrightarrow{x}$
satisfies the distribution with the probability density $P_{GM}(\overrightarrow{x})$:

\begin{equation}
P_{GM}(\overrightarrow{x})=\sum_{n=1}^{N}A_{i}P(\overrightarrow{x}|\theta_{i})
\label{eq:GMclusters}
\end{equation}

\begin{equation}
\sum_{n=1}^{N}A_{i}=1
\label{eq:GMclusterWeights}
\end{equation}
Here $P(\overrightarrow{x}|\theta_{i})$ is the normal distribution
described by the set of its unknown parameters $\theta_{i}$ (mean
and covariance matrix), and unknown weights $A_{i}$ of the corresponding
distribution in the sum. N is the number of distributions in the mixture.
The optimal distributions $P_{GM}(\overrightarrow{x})$ and $P(\overrightarrow{x}|\theta_{i})$,
as well as $N=N_{BIC},A_{i},\theta_{i}$, at which the minimum of
the BIC criterion is reached, will be the result of this stage.

\begin{figure}
\centering
\includegraphics[scale=0.5]{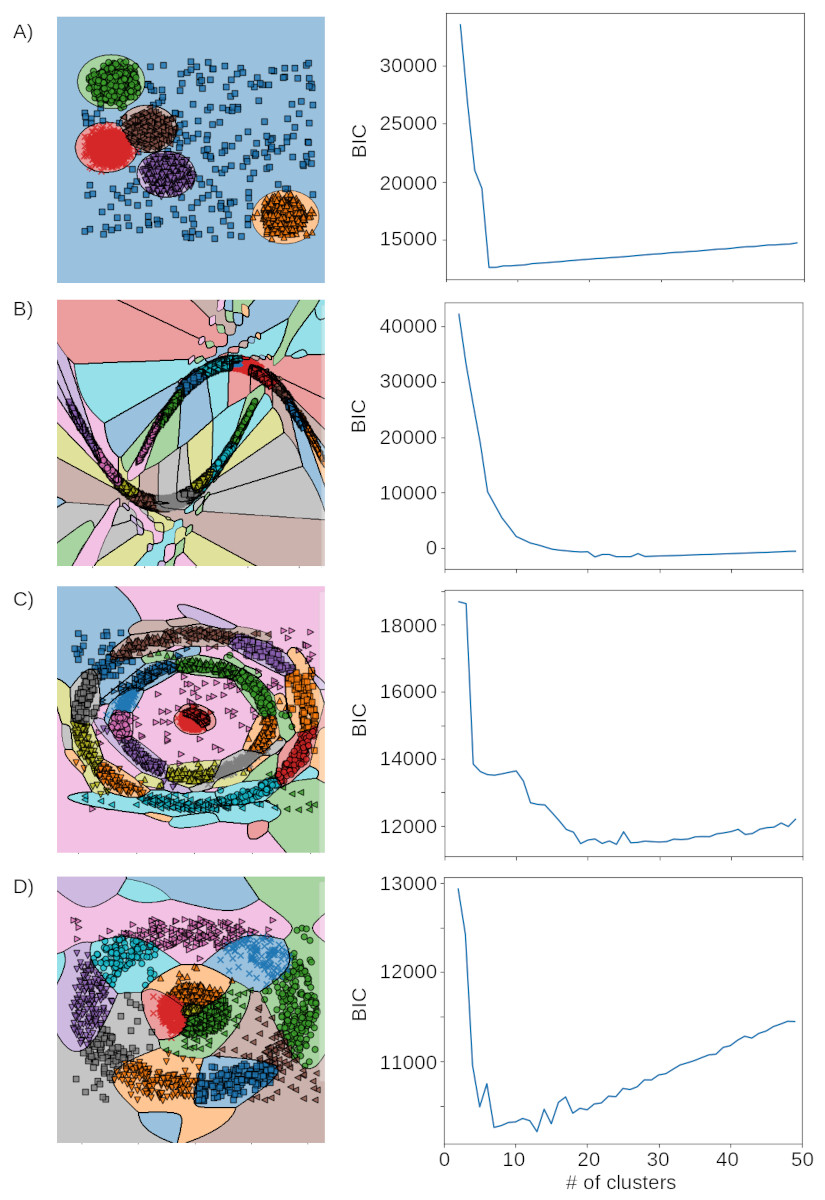} 
\caption{Clustering points into BIC-optimal mixture of Gaussian distributions.
On the left is a separation by the number of clusters $N_{BIC}$ corresponding
to the BIC minimum. The background corresponds to decision areas,
all points of which the algorithm will refer to given cluster. Colors
correspond to different clusters. On the right is the dependence of
BIC on the number of clusters. A) 'medium blobs'; B) 'two horseshoes';
C) 'three nested rings with noise'; D) 'two nested rings with noise'}
\label{fig:BICclustering} 
\end{figure}

Figure~\ref{fig:BICclustering} shows examples of the BIC dependence
on the number of clusters $N$ and the resulting splitting of the
original data set into the optimal number $N_{BIC}$ of Gaussian clusters
$P(\overrightarrow{x}|\theta_{i})$, as well as decision regions,
defining the boundaries between these clusters. Obviously, only the
first example ('blobs', shown Figure~\ref{fig:BICclustering}A) is an
adequate enough model; the other cases (Figure~\ref{fig:BICclustering}B-D)
are just some clusterings with high enough PWTN.

\subsection{Stage 2: Calculation of intercluster distances}

We require the algorithm to be robust to a noise and interpretable
from statistical point of view. So metrics should be carried out not
in Euclidean space, but in probability space. We used a mixture of
Gaussian distributions so the distance between clusters and cluster
sizes are calculated using Mahalanobis distances \cite{Mahalanobis},
native for gaussian statistical distribution:

\begin{equation}
p_{ij}=\sqrt{(\overrightarrow{x_{i}}-\overrightarrow{y_{j}})S_{j}^{-1}(\overrightarrow{x_i}-\overrightarrow{y_j})^{T}}
\label{eq:MahalanobisD}
\end{equation}
where $S_{j}^{-1}$ is the matrix inverse to the covariance matrix
of the cluster to which point $\overrightarrow{y_j}$ belongs.

The distance is widely used in different clustering and classification
tasks \cite{MNClust,McLachlan_1992}. Qualitatively, the Mahalanobis distance is
the distance from a point $\overrightarrow{x}$ to an ellipsoid normalized
to the ellipsoid width. Therefore it could be qualitatively interpreted
as t-statistic value\cite{Student_1908}. Thus, the Mahalanobis distance
can be used as a statistic for testing the hypothesis that the point
$\overrightarrow{x}$ belongs to a given normal distribution, to which
$\overrightarrow{y}$ belongs. When it reaches a certain threshold
value, we can talk about the rejection of this
hypothesis with the corresponding statistical significance level.

The Mahalanobis distance is usually
defined for the case when both points are from the same distribution.
We will use points from different distributions (clusters). Accordingly,
their covariance matrices ($S_{j}$ and $S_{i}$) in the general case
will be different, and the distance calculated (\ref{eq:MahalanobisD})
will not be a distance, since it is not symmetrical:

%% \sqrt{\overrightarrow{x}S_{y}^{-1}\overrightarrow{y}^{T}}

\begin{equation}
\sqrt{(\overrightarrow{x_i}-\overrightarrow{y_j})S_{j}^{-1}(\overrightarrow{x_i}-\overrightarrow{y_j})^{T}}\ne
 \sqrt{(\overrightarrow{x_i}-\overrightarrow{y_j})S_{i}^{-1}(\overrightarrow{x_i}-\overrightarrow{y_j})^{T}}
\label{eq:MahalanobisDunsym}
\end{equation}

In this case, when analyzing points from two clusters, 
we have 4 statistical distributions of these quasi-metrics:
$p_{ii},p_{jj},p_{ji},p_{ij}$, where the indices i,j
correspond to the points in i-th and j-th clusters, obtained in Stage
1 GM/BIC clusterization (\ref{eq:GMclusters}). The distributions of
$p_{ii},p_{jj}$ can be used for calculating the cluster sizes, and of $p_{ij},p_{ji}$
- for calculating intercluster distances with taking into account the j-th and i-th
cluster shapes, respectively.

The intercluster distance $R_{ij}$ must be symmetrical ($R_{ij}=R_{ji}$),
so we should form a symmetric function from $p_{ij},p_{ji}$ distributions.
As the intercluster distance we choose the maximum between 5-th percentiles:

\begin{equation}
\begin{array}{c}
R_{ij}=Max(Percentile_{5\%}(p_{ij}),Percentile_{5\%}(p_{ji}))\\
R_{ii}=0
\end{array}
\label{eq:InterclusterDistances}
\end{equation}

The 5th percentile (5\%) was chosen for statistical reasons and corresponds
to the lower limit, below which, with a standard statistical confidence
level of 95\%, the distance between points in these clusters does
not fall down. In addition, percentiles are more robust to outliers
than standard deviations and means. The maximization between two values
is used to symmetrize the distance $R_{ij}$. It corresponds to the transition
to the coordinate system of the cluster in which the intercluster
distance is higher, and the two clusters can be more confidently separated.

\begin{figure}
\centering
\includegraphics[scale=0.5]{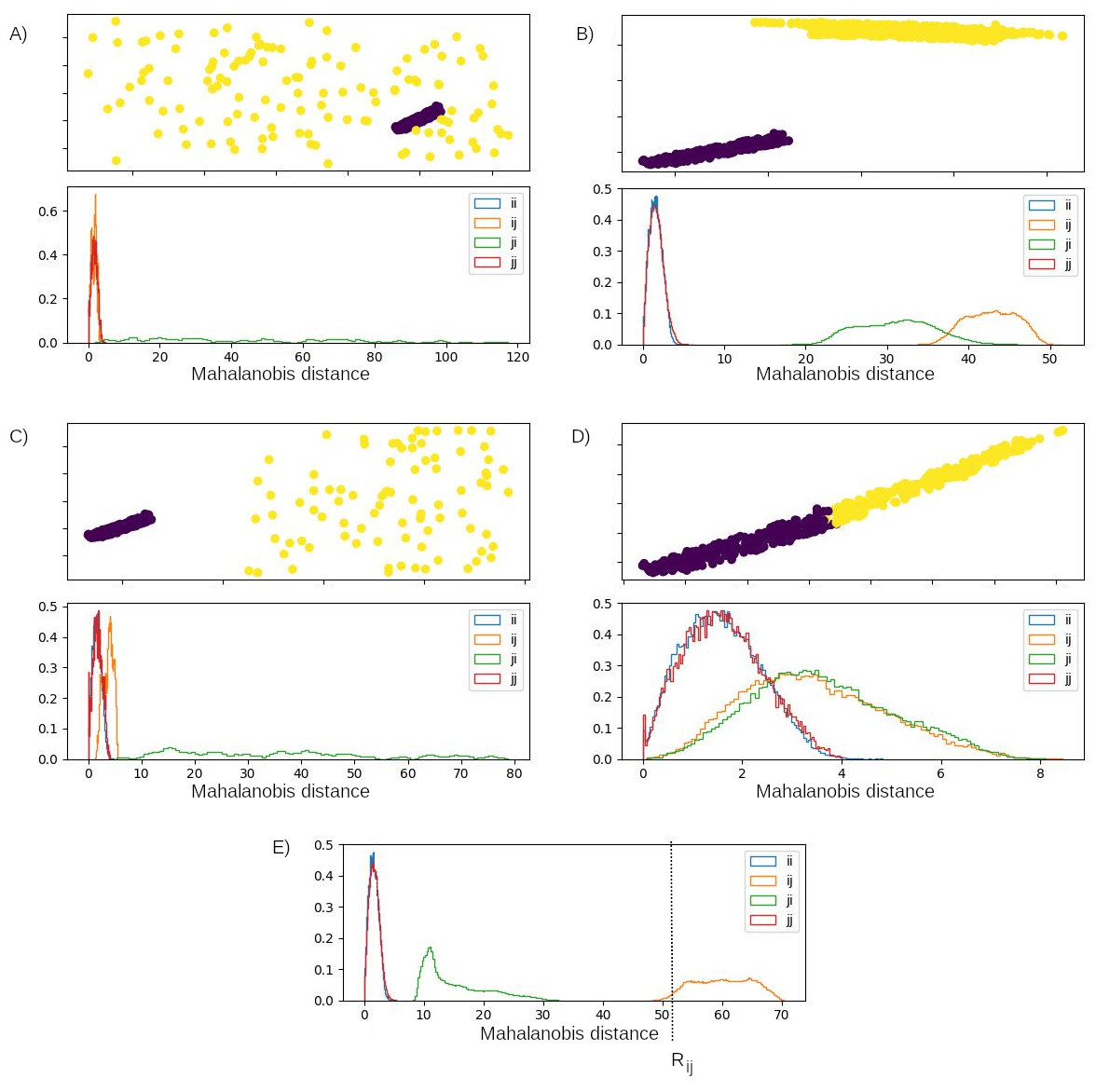} 
\caption{A-D) Distributions of points within clusters (different colors correspond
to different clusters) and distributions of Mahalanobis distances
corresponding to them (orange and green correspond to distributions
of intercluster distances $p_{ij},p_{ji}$, red and blue correspond
to cluster size distributions $p_{ii},p_{jj}$). E) An illustration
of determining the intercluster distance
$R_{ij}$ from the distributions of Mahalanobis distances for two
clusters. }
\label{fig:MahalanobisDistanceDistribution} 
\end{figure}

Figure~\ref{fig:MahalanobisDistanceDistribution}A-D shows examples of
clusters and their corresponding Mahalanobis distance distributions
(\ref{eq:MahalanobisD}). It can be seen from the figure that the
distance distributions $p_{ij}$ and $p_{ji}$ generally do not coincide,
which explains distance symmetrization (\ref{eq:InterclusterDistances}).
The scheme for determining distances is illustrated in Figure~\ref{fig:MahalanobisDistanceDistribution}E.

\subsection{Stage 3: Grouping clusters into superclusters}

DBSCAN was used as an algorithm for grouping the clusters into superclusters,
matrix of intercluster distances $R_{ij}$ \ref{eq:InterclusterDistances}
is used as inter-object distances. To simplify the enumeration of
the iterations over the maximal element distance in the supercluster
$\varepsilon$, the DBSCAN $\varepsilon$ hyperparameter is used as
iteration identifier. To do this all unique values of $\varepsilon$
were ordered in increasing order, and intermediate values 

\begin{equation}
\overline{\varepsilon}_{k}=(\varepsilon_{k-1}+\varepsilon_{k})/2;\varepsilon_{0}=0
\label{eq:vareps}
\end{equation}
between the neighbors of this sequence were calculated. The resulting
sequence of $\overline{\varepsilon}_{k}$ is used as iteration indices.
Obviously, at the smallest distance $\overline{\varepsilon}_{1}$
DBSCAN produces the maximum possible value of clusters $N_{BIC}$
(determined from the optimality of the BIC criterion at the first
stage of the algorithm), and at the largest $\overline{\varepsilon}_{k_{max}}$
DBSCAN produces only one cluster. In order to speed up algorithm,
if iterating over 10 consecutive values of $\overline{\varepsilon}_{k}$
DBSCAN produces the number of clusters equal to 1, the iteration process
over $\overline{\varepsilon}_{k}$ is stopped.

Further calculations of stop criteria require a matrix of distances
between new superclusters $D_{ij}(\overline{\varepsilon})$. The distance
$D_{SC_{i},SC_{j}}(\overline{\varepsilon})$ between two superclusters
$SC_{i},SC_{j}$ is taken to be the minimum distance between clusters
$C_{n},C_{m}$ belonging to these two superclusters :

\begin{equation}
D_{SC_{i},SC_{j}}(\overline{\varepsilon})=Min_{m,n:C_{m}\subseteq SC_{i},C_{n}\subseteq SC_{j}}(R_{mn})
\label{eq:D_inter}
\end{equation}

In terms of speed, this algorithm is intermediate between a greedy
algorithm (for example, agglomerative) and an exhaustive search. At
the same time, at each iteration of $\overline{\varepsilon}$, it
accurately estimates the number of superclusters produced by this
solution. It should be noted that despite the fact that the DBSCAN
method is a metric method, the used distance matrix $R_{ij}$ (defined
in the previous step) is essentially a set of t-statistic values.
Therefore, when performing this stage, we cluster the differences
in the statistical characteristics of clusters, rather than their
Cartesian characteristics.

\subsection{Stage 4: Stop criteria}

An important task is the choice of the optimal superclusters configuration
and their number, which is reduced to the choice of the optimality
criterion. It is obvious that new use of the BIC criterion will not
bring any sufficient results: according to this criterion, we have
already chosen the maximum number of clusters $N_{BIC}$. The other
widely used criterion is the Silhouette criterion. However, it has
a problem - it is related with the concept of subtracting distances
(and hence connected with the Euclidean metric), which makes its statistical
interpretation difficult, when distances are t-statistics. Therefore,
we introduce a more clear criterion from statistical point of view - the
matrix quality criterion (MC).

Let us formulate the problem of proximity of two clusters in terms of testing hypothesis.
The null hypothesis is that $i,j$ clusters are close and the minimal distance 
$D_{ij}$ between them  is statistically insignificant and statistically corresponds to 
the distances within single cluster.
Alternative hypothesis that they are far from each other.
Let us use the intercluster distance $D_{ij}$ as statistics for the testing this hypothesis.
The distribution of distances $p_{ii}$ between points within a single gaussian cluster is the 
distribution of this statistics when null hypothesis is true. 

One-sided $\alpha$-level statistical test for this criteria will be: 
\begin{equation}
 p_{value}(D_{ij})<\alpha
\label{eq:hypotest_rule}
\end{equation}
where $p_{value}$ is calculated over null-hypothesis distribution (i.e. over $p_{ii}$ distribution)

For  Mahalanobis distance $p_{ii}$, the $p_{ii}^2/2$ value has $\chi^2$ distribution:
\begin{equation}
p_{ii}^{2}/2\sim\chi^{2}(N_X)
\label{eq:D_dist}
\end{equation}
where degrees of freedom $N_X$ is the 
dimension of the clustered points.

Due to for one-sided test $p_{value}(D)$ decreases with D increasing, 
the null hypothesis rejecting rule (\ref{eq:hypotest_rule}) can be rewriten in form:

\begin{equation}
D_{ij}(\overline{\varepsilon}_{k})>\delta D_{\alpha}
\label{eq:hyptest}
\end{equation}
where 
\begin{equation}
\delta D_{\alpha}=\sqrt{2\cdot Q_{1-\alpha}}
\label{eq:D_chi2}
\end{equation}
following (\ref{eq:D_dist}), and calculated from quantile value $Q_{p}$ of $\chi^{2}(N_X)$ distribution with
given significance level $\alpha$ and known degrees of freedom $N_X$.

Following this rule the superclusters i-th and j-th are close (the null hypothesis can not be rejected) 
when condition (\ref{eq:hyptest}) is not met. In this case the number of superclusters
can be reduced by making next iteration $\overline{\varepsilon}_{k+1}$.

This hypothesis testing rule (\ref{eq:hypotest_rule}) can be interpreted in terms of Mahalanobis distance (\ref{eq:hyptest})
by the following: if two superclusters are located at a distance $D_{ij}(\overline{\varepsilon})$
less than $\delta D_{\alpha}$, they can be considered close
with a significance level $\alpha$. Therefore, as a matrix quality
criterion (MC), we use the proportion of superclusters that do not
have close superclusters:

\begin{equation}
MC(\overline{\varepsilon})=\frac{\sum_{i\in[1..N_S]}[Min_{j,j\neq i}(D_{ij}(\overline{\varepsilon}))>\delta D_{\alpha}]}{N_S(\overline{\varepsilon})}
\label{eq:MCcriterium}
\end{equation}
where [x] is 1 when condition x is met, 0 otherwise, $N_S(\overline{\varepsilon})$ is a current number of superclusters  (dimension of $D_{ij}$)
at stage $\overline{\varepsilon}$.

For each supercluster, we determine whether the supercluster closest
to it is separable from it or not. We divide the resulting number
of separable superclusters by the total number of superclusters.

If we transfer all the distances $D_{ij}$ to p-values using (\ref{eq:D_chi2}),
the MC criterion will become:

\begin{equation}
MC(\overline{\varepsilon})=\frac{\sum_{i\in[1..N_S]}[Max_{j\neq i}(p_{value}(D_{ij}(\overline{\varepsilon})))<\alpha]}{N_S(\overline{\varepsilon})}
\label{eq:MCcriterium_pval}
\end{equation}
and has clear statistical interpretation: maximal value 1 of this
criterion demonstrates that every supercluster at stage $\overline{\varepsilon}$ is separated from others
with significance level at least $\alpha$. 

Obviously, $MC(\overline{\varepsilon})$ lies within $[0..1]$, and
reaches its maximum value of 1 when we find the clustering 
where all superclusters are statistically
significant separable from each other with significance level $\alpha$.
The MC criteria do not depend on cluster sizes due to the sizes 
are similar and theoreticaly predicted from $\chi^2$ distribution (\ref{eq:D_dist}).

\begin{figure}
\centering
\includegraphics[scale=0.38]{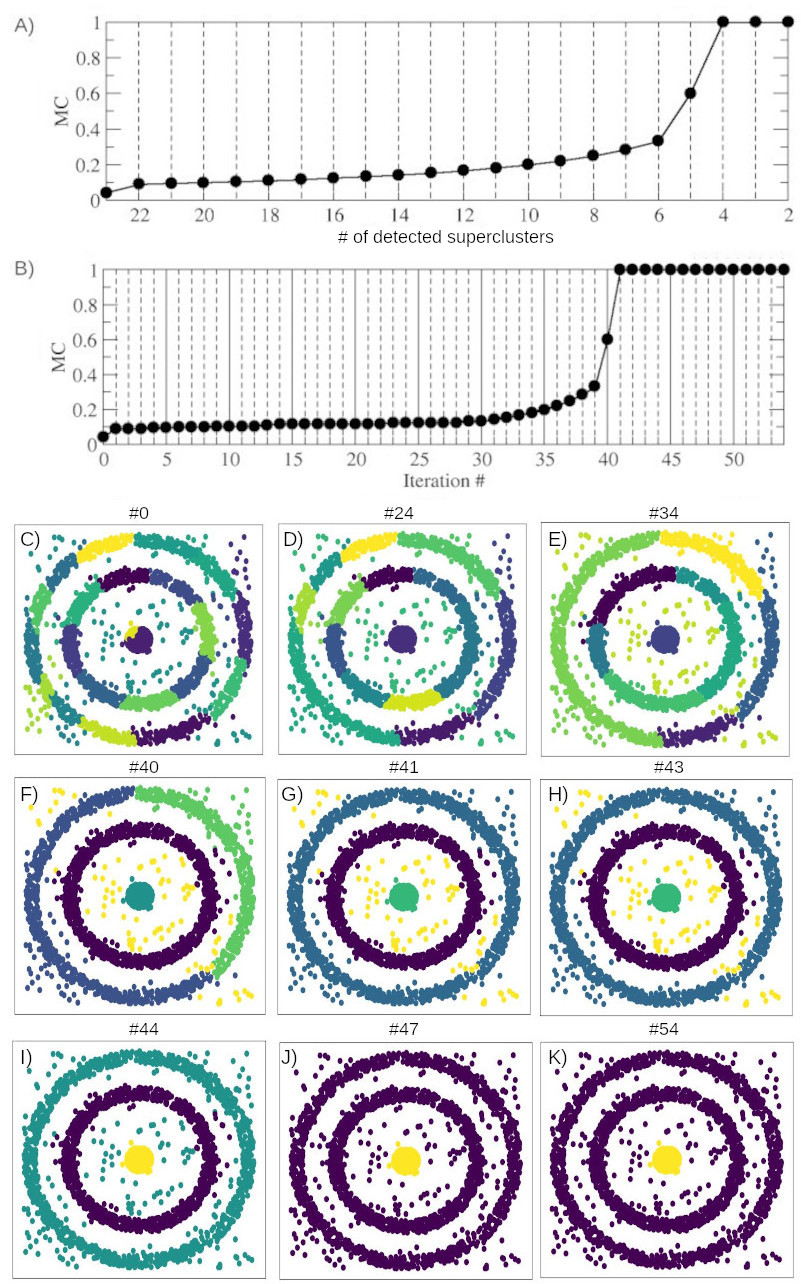} 
\caption{Dependence of the MC (A), number of superclusters (B) and superclustering
during training (C-K) as function of iteration number using 'three
nested rings with noise' data set as an example for $\alpha=0.1$.}
\label{fig:MCmetricSequence} 
\end{figure}

Figure~\ref{fig:MCmetricSequence}A-B shows dependence of MC criteria
on iteration number and number of superclusters. Figure~\ref{fig:MCmetricSequence}C-J
shows the shapes of superclusters tested during training. The Figure~\ref{fig:MCmetricSequence}A-B
demonstrates the need to stop at the $\overline{\varepsilon}$ stage,
at which $MC(\overline{\varepsilon})$ becomes 1 : after this the
number of superclusters only decreases, and the value of the $MC(\overline{\varepsilon})$
criterion does not change. From the example shown in the figure, it
can be seen that for the first time MC reaches the value of 1 at the
41th iteration, which leads to the splitting of the data set into 4
superclusters we expected - three nested rings and a noise. Further
iterations do not improve the clustering, and leave MC unchanged.

\section{Final algorithm}
The final GMSDB algorithm consists of the following actions:
\begin{enumerate}
\item Get data set of $N_X$-dimension points $\overrightarrow{X}$;
\item For each number of clusters $N$ in given range (from 2 to a large enough number) make Gaussian Mixture clusterization of $\overrightarrow{X}$ 
- find parameters in eq.(\ref{eq:GMclusters});
\item Find number of clusters $N_{BIC}$ for which the BIC is minimal and use corresponding optimal GM clusterization ($C_m$)
of $\overrightarrow{X}$ for the following steps;
\item Calculate the cluster distance matrix $R_{mn}$ using each pair of found clusters $C_m,C_n$, 
all their points pairs:
\[
 \overrightarrow{x_m},\overrightarrow{y_n}\in \overrightarrow{X}: \overrightarrow{x_m}\in C_m, \overrightarrow{y_n}\in C_n
\],
eqs.(\ref{eq:MahalanobisD},
\ref{eq:InterclusterDistances}), and covariance matrices $S_m,S_n$ for these clusters obtained at the previous step;
\item Get ordered sequence of unique distances $\varepsilon_{k} \in R_{mn}, m\ne n, \varepsilon_{k}>0$, calculate $\overline{\varepsilon}_{k}$ 
using eq.(\ref{eq:vareps});
\item For each $\overline{\varepsilon}_{k}$ from its smallest to its largest value do the following:
\begin{enumerate}
 \item Cluster the objects (clusters $C_m$), defined by the distance matrix $R_{mn}$ 
into the groups (superclusters $SC_j$) by DBSCAN method
with its hyperparameter $\varepsilon=\overline{\varepsilon}_{k}$, 
get the number of found superclusters $N_S$;
 \item Calculate the supercluster distance matrix $D_{ij}$ between found superclusters $SC_i,SC_j$ using eq.(\ref{eq:D_inter}) 
and cluster distance matrix $R_{mn}$, calculated at step 4;
 \item Calculate MC from $D_{ij}$ using eq.(\ref{eq:MCcriterium},\ref{eq:D_chi2}), 
given data dimension $N_X$, number of found superclusters $N_S$, and given significance level $\alpha$ (default value $\alpha=0.1$);
 \item If MC is equal to 1 - exit the cycle;
\end{enumerate}
\item Return found superclusters $SC_j$ as the solution.
\end{enumerate}

\section{Experiments}

At first we test the algorithm in noise-free situations. The algorithm
was tested on artificial data sets, the code of which is given in \cite{GMSDB}.
Some of these data sets are identical to those used for sklearn library
\cite{SKLEARN} clustering tests (https://scikit-learn.org/stable/modules/clustering.html).
All the data sets were generated by various transformations from a
normal and uniform distributions, so data sets have varying cluster
shapes and point densities within clusters.

\begin{figure}
\centering
\includegraphics[scale=0.5]{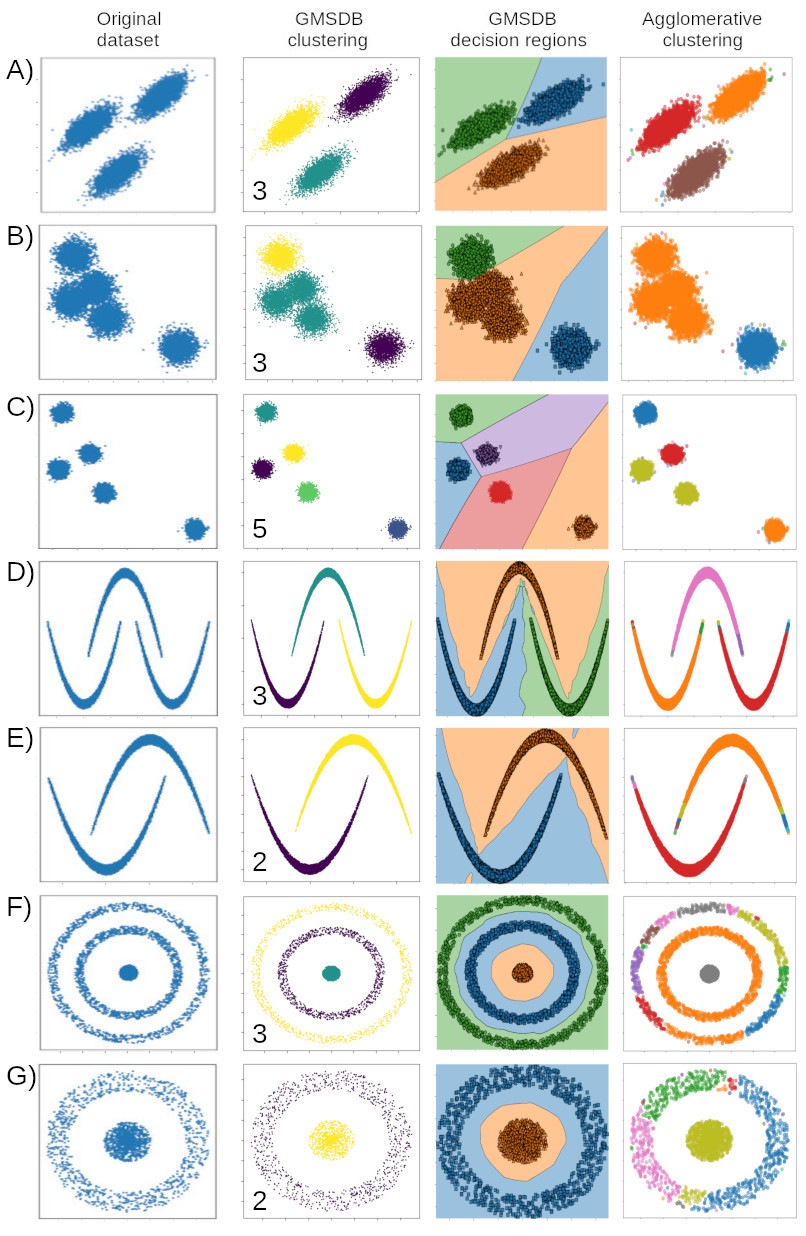} 
\caption{Comparison of clustering of different identical data sets by GMSDB
and agglomerative algorithms (minimal interobject distance as intercluster
distance, metric - euclidean). Noise-free situation. The maximum allowed
number of clusters is 50, $\alpha=0.1$. From left to right - the original data set,
clustering by the GMSDB algorithm and the found optimal number of
superclusters, decision regions of the trained GMSDB algorithm, clustering
by the agglomerative algorithm. Bottom left corner in second column
images - the number of superclusters detected by GMSDB.}
\label{fig:NoiseFreeClusteringExamples} 
\end{figure}

Figure~\ref{fig:NoiseFreeClusteringExamples} shows the results of such
testing. The 'three grains' (Figure~\ref{fig:NoiseFreeClusteringExamples}A),
large (Figure~\ref{fig:NoiseFreeClusteringExamples}B) and small (Figure~\ref{fig:NoiseFreeClusteringExamples}C)
'blobs', which are useful for testing clustering with statistical
models and models based on centroids and distances between them; 'three
horseshoes' (Figure~\ref{fig:NoiseFreeClusteringExamples}D) and 'two
horseshoes' (Figure~\ref{fig:NoiseFreeClusteringExamples}E), which are
useful for testing clustering with distance-based models, and 'three
nested rings' (Figure~\ref{fig:NoiseFreeClusteringExamples}F) and 'two
nested rings' (Figure~\ref{fig:NoiseFreeClusteringExamples}G).

The results of the GMSDB algorithm and the Agglomerative algorithm
\cite{Augmentation} was compared in the following way. The maximum
possible number of clusters is 35, which exceeds $N_{BIC}$ for any
data set. For agglomerative clustering, the distance is Euclidean,
and the distance between clusters is calculated by their nearest elements.

The figure shows that in the case of well-separated 'grains' (Figure~\ref{fig:NoiseFreeClusteringExamples}A)
and 'small blobs' (Figure~\ref{fig:NoiseFreeClusteringExamples}C) the
algorithms work very similarly and the results are the same as expected
- 3 and 5 main classes correspondingly. It can be seen that the agglomerative
algorithm leaves artifacts on the cluster boundaries. The GMSDB algorithm
produces no artifacts, and accurately guesses the number of superclusters.
A similar situation occurs in the case of 'three horseshoes' and 'two
horseshoes' (Figure~\ref{fig:NoiseFreeClusteringExamples}D,E) - agglomerative
clustering guesses the main body of the cluster quite well, but the
edges with clustering artifacts become even larger. The GMSDB algorithm
continues to correctly guess the number and shape of superclusters.
'Big blobs' (Figure~\ref{fig:NoiseFreeClusteringExamples}B), where the
distance between clusters is of the order of their size, is problematic
for both algorithms, but GMSDB guesses a little closer than the agglomerative
one. The biggest difference is in 'three nested rings' and 'two nested
rings' (Figure~\ref{fig:NoiseFreeClusteringExamples}F,G) - the agglomerative
one cannot adequately separate them - as the radius increases (and
the density of points decreases), the ring is divided into an increasing
number of clusters. Unlike the agglomerative algorithm, GMSDB adequately
clusterizes these data sets.

\begin{figure}
\centering
\includegraphics[scale=0.55]{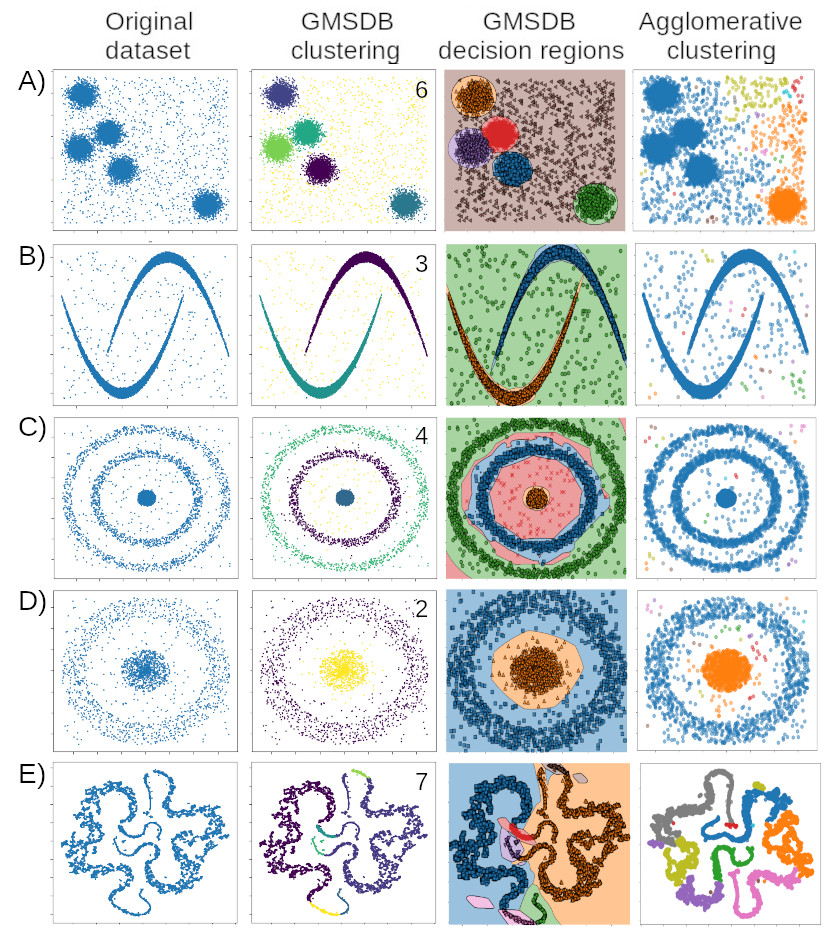} \caption{Comparison of clustering of different identical data sets by GMSDB
and agglomerative algorithms (minimal interobject distance as intercluster
distance, metric - euclidean). Noisy situation. The maximum allowed
number of clusters is 50, $\alpha=0.1$. From left to right - the original data set,
superclustering by the GMSDB algorithm and the found optimal number
of superclusters, decision regions of the trained GMSDB algorithm,
clustering by the agglomerative algorithm. Top right corner in second
column images - the number of superclusters detected by GMSDB.}
\label{fig:NoisyClusteringExamples} 
\end{figure}

Examples of testing at noisy data sets are shown in Figure~\ref{fig:NoisyClusteringExamples}.
Here the situation is much different. The impact of noise on 'middle
blobs' clustering (Figure~\ref{fig:NoisyClusteringExamples}A) makes
classification more difficult with the agglomerative method, but still
allows clustering with the GMSDB algorithm.

The last test is 'two snakes' (Figure~\ref{fig:NoisyClusteringExamples}E)
- the result of the t-SNE transform \cite{TSNE} of the 'two noisy
horseshoes' data set (Figure~\ref{fig:NoisyClusteringExamples}B). Two
snakes - a data set of two clusters and noise, which has a complex
shape and an inhomogeneous noise-like structure inside and outside
clusters. The distances between two clusters in different regions
are different and comparable with the characteristic sizes of the
clusters themselves. This data set cannot be adequately clustered by
both algorithms (agglomerative and GMSDB), but it can be easily seen
that the GMSDB algorithm finds fewer superclusters than the
agglomerative algorithm and detects two largest superclusters corresponding
to 'snakes' sufficiently well.

\section{Discussion}

According to Kleinberg's impossibility theorem \cite{Kleinberg_2002},
there is no clustering algorithm that is simultaneously scale-invariant,
consistent, and rich. Obviously, the proposed GMSDB algorithm is unrich
- it does not enumerate all clustering/superclustering variants. Let
us evaluate how it differs from the rich one in order to understand
what possible clusterings it is limited to. Obviously, if the number
of Gaussian distributions is equal to the number of points in the
data set, then the algorithm 'find the parameters of all Gaussian distributions,
and then combine them to obtain a clustering labeled by expert' is
almost rich - for discrete points (we usually work with such data sets),
it is enough to choose Gaussian distributions much narrower than the
minimum distance between points. Compared to the full algorithm in
the GMSDB algorithm we limit the number of normal distributions and
their shape: their shape is determined from the EM algorithm by the
maximum likelihood condition, and the number is determined from the
minimum of the BIC criterion. The possible combinations of clusters
(superclusters) in the GMSDB algorithm are also limited - there should
be the maximum number of superclusters separable by the Mahalanobis
distance with a significance level $\alpha$. This limits the set
of possible clusterings that the algorithm is able to determine adequately.
In practice, it is not able to correctly separate non-connected or
significantly overlapping clusters, as well as clusters in which there
are not enough elements for the EM algorithm to work or which are
'not likelihood enough', and could be clustered by different way with
more likelihood in terms of Gaussian Mixture.

The high quality of the algorithm is demonstrated in noiseless (Figure~\ref{fig:NoiseFreeClusteringExamples})
and noisy (Figure~\ref{fig:NoisyClusteringExamples}) situations compared
to the agglomerative algorithm.

The advantages of the GMSDB method include also its statistical formulation,
which makes it possible to estimate the probabilities of points belonging
to a particular supercluster, and use it to classify new data both
in the hard clustering mode and in the soft (fuzzy) clustering mode.
Indeed, it is known that a mixture of Gaussians is capable to produce
soft clustering - to estimate for each point the probability of its
belonging $P_{C_{j}}(\overrightarrow{x})$ to one or another Gaussian
cluster $C_{j}$. Each supercluster $SC_{i}$ is a union of the original
Gaussian clusters and the sets of clusters in different superclusters
do not intersect. So the probabilities of any new point belonging
to supercluster $P_{SC_{i}}(\overrightarrow{x})$ can be calculated
from the probabilities of belonging this point to the clusters $C_{j}\subseteq SC_{i}$
that form this supercluster:

\begin{equation}
P_{SC_{i}}(\overrightarrow{x})=\sum_{j:C_{j}\subseteq SC_{i}}P_{C_{j}}(\overrightarrow{x})\label{eq:SuperCLProbab}
\end{equation}

Table \ref{tab:time} shows the average time of clustering data sets by the
GMSDB and Agglomerative algorithms (in seconds, by an identical computer),
as well as the Rand Index (RI) confidential interval over 10 runs, determined from the initial labeling
of the test data sets during their creation. In noisy cases, noise
was marked as a separate class.

From Table \ref{tab:time} it can be seen that the GMSDB algorithm
is most often more adequate than the agglomerative algorithm, especially
on noise-free data sets. Higher RI values indicate higher adequacy.

The main disadvantage of the algorithm is its speed, shown in Table
\ref{tab:time}. It can be seen from the table that GMSDB is two orders
of magnitude slower than the agglomerative algorithm, so its use is
recommended when one takes into account its low speed, in the case
when the adequacy of clustering is very important, and there is much
enough data for its operation. The algorithm is implemented in Python,
so using fast search algorithms or other programming languages for
these tasks could improve its performance. Its current variant \cite{GMSDB}
has implementations for the speed optimization by a kind of dichotomy
search for stage 1 and Monte-Carlo search for stage 3, significantly
improving clustering speed.

\begin{table}
\caption{The execution time for clustering identical data sets by the GMSDB
and Agglomerative algorithms (in seconds), as well as the Rand Index
(RI) determined from the initial labeling of test data sets during
their creation (in noisy cases, noise is marked as a separate class).
Bold indicates the best RI result for given data set.}
\label{tab:time} 

%% \begin{tabular}{|p{2cm}|p{1cm}|p{1cm}|p{1cm}|p{1cm}|p{1cm}|p{1cm}|p{1cm}|}
\begin{tabular}[t]{|c|c|c|c|c|c|c|c|}
\hline
& \multicolumn{5}{|c|}{GMSDB}& \multicolumn{2}{|c|}{Agglomerative} \tabularnewline
\hline
data set  & St.1  & St.2 & St.3,4  & Tot.time & RI interval (0.95) & time  & RI interval (0.95) \tabularnewline
\hline
\multicolumn{8}{c}{Noise-free data sets (Figure~\ref{fig:NoiseFreeClusteringExamples})}\tabularnewline
\hline
Grains &  10.0 & 21.7 & 0.02 & 32.0 & \bf{1.0} & 0.67 & 0.998 \tabularnewline
Big blobs &  9.7 & 14.6 & 0.01 & 24.3 & \bf{0.76} & 0.62 & 0.52 \tabularnewline
Small blobs &  11.1  & 14.4 & 0.01  & 25.5 & \bf{1.0} & 0.60 & 0.999 \tabularnewline
3 horseshoes &  17.6 & 44.0 & 0.04  & 61.6 & \bf{1.0} & 1.6 & 0.987 \tabularnewline
2 horseshoes &  19.5 & 20.0 & 0.02 & 39.5  & \bf{1.0} & 0.68 & 0.975 \tabularnewline
3 nested rings &   3.4  &  3.5  &  0.04  &  6.9  & \bf{1.0}  &  0.08  & 0.9 \tabularnewline
2 nested rings &  1.0 & 0.7 & 0.04 & 1.7 & \bf{1.0} & 0.046 & 0.825 \tabularnewline

\hline
\multicolumn{8}{c}{Noisy data sets (Figure~\ref{fig:NoisyClusteringExamples})}\tabularnewline
\hline
Medium blobs &  10.6 & 21.3 & 0.02 & 31.9 & \bf{0.985..0.988} & 0.78 & 0.18..0.51  \tabularnewline
2 horseshoes &  17.6 & 21.2 & 0.1 & 38.8 & \bf{0.994..0.996} & 0.66 & 0.48..0.49 \tabularnewline
3 nested rings &   5.6  &  3.4  &  0.1  &  9.1  & \bf{0.942..0.955} &  0.09  & 0.30..0.72 \tabularnewline
2 nested rings &  2.6 & 1.2 & 0.01 & 3.8 & 0.880..0.883 & 0.059 & 0.42..0.89 \tabularnewline
2 snakes &  17.5 & 19.2 & 0.4 & 37.1 &  0.72..0.83 & 0.71 & 0.65..0.85 \tabularnewline
\hline

\end{tabular}
\end{table}

The next drawback is the presence of a control hyperparameter - the
significance level $\alpha$. To obtain all the results described
above, we used $\alpha=0.1$, exceeding standard statistical value 0.05.

\begin{figure}
\centering
\includegraphics[scale=0.35]{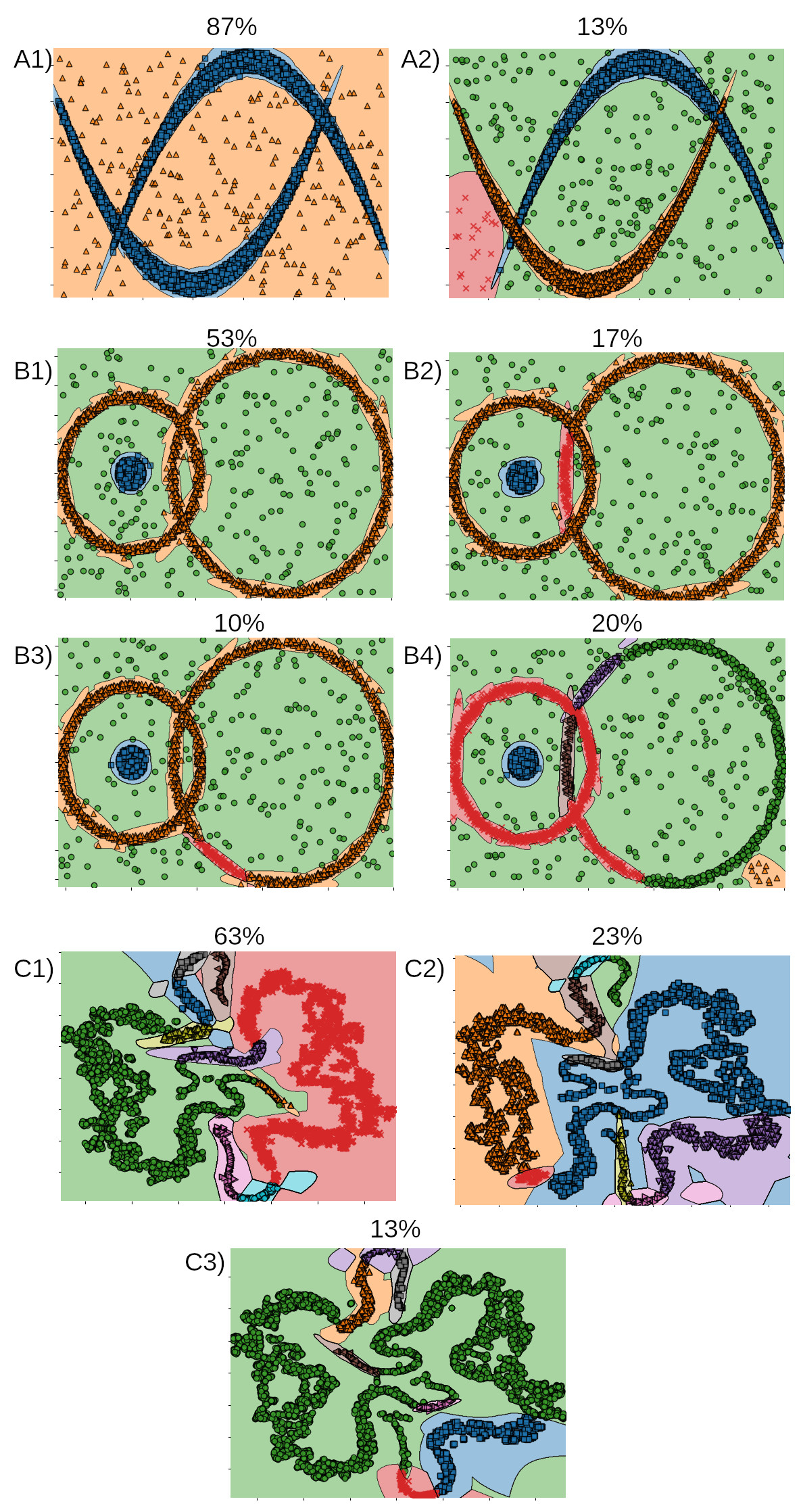} \caption{The effect of stochasticity to clustering in the case of intersecting
clusters and the probability of occurrence of these cases over 30 runs (top). 
A1-A2) two intersecting horseshoes, B1-B4) intersecting rings, C1-C3) two snakes }
\label{fig:CrossingClusters} 
\end{figure}

The last disadvantage of the method is the stochasticity of the results
- the optimal separation of the same data set can change from run to
run. This is due to the algorithm is based on the EM implementation
of GM, which is based on an iterative algorithm and depends on the
initial conditions. Examples and their probabilities
are shown in Figure~\ref{fig:CrossingClusters}.

\section{Conclusion}

The paper presents a statistical algorithm for GMSDB adequate clustering
a data set by an optimal grouping of inadequate gaussian models.

The algorithm performs two sequential operations - approximates the
set of points under study by the optimal superposition of normal distributions,
each of them, taken separately, is generally not an adequate cluster
in the data. To create a model of adequate clusters, the algorithm
groups the resulting Gaussian distributions into the maximum number
of separable superclusters (with given statistical significance level $\alpha=0.1$).
Thus, the algorithm clusters the data set into statistically separated
superclusters, each of which is described by a superposition of normal
distributions.

The algorithm consists of three main stages: approximates the
data set by a mixture of Gaussian distributions, the number of which
is determined based on the minimum of the BIC criterion; uses the
Mahalanobis distance for estimating the statistical distances between
clusters and cluster sizes; combines the resulting clusters into superclusters
using the DBSCAN method by sequential iterations over its hyperparameter
(maximum distance $\varepsilon$) from smaller to larger values; stops
the iterations over the hyperparameter when the matrix quality criterion
MC introduced by us reaches its maximum value of 1. The matrix quality
criterion calculates the proportion of statistically significant separable
superclusters among all found superclusters.
Therefore the algorithm joins hierarchical and partitional methods into a single 
scheme based on statistical theory testing approach. 

Although the algorithm is computationally very slow, it shows good
results on several test data sets, in noise and noiseless situations. 
An essential advantages of the method
are: the ability to predict a cluster for new data based on an already
trained clusterer, and the possibility of soft (fuzzy) clustering.
The Python source code of the clusterer is available in \cite{GMSDB} 
and available at PyPI as gmsdb package.

%% Here is a citation \cite{chow:68}.

\section*{Acknowledgements}
The work has been done under financial support of the Ministry of
Science and Higher Education of the Russian Federation (Subsidy No.075-GZ\/C3569\/278).

\bibliographystyle{abbrv}

\bibliography{biblio}

\end{document}